# Enhanced Fault Detection and Cause Identification Using Integrated Attention Mechanism

Mohammad Ali Labbaf Khaniki*, Alireza Golkarieh**, Houman Nouri***, Mohammad Manthouri****

*Faculty of Electrical Engineering, K.N. Toosi University of Technology, Tehran, Iran

**University of Michigan, Department of Mechanical Engineering

***Data science, Mathematics department, university of Sussex, Brighton, un

**** Faculty of Electrical and Electronic Engineering, Shahed university, Tehran, Iran

mohamad95labafkh@gmail.com*

Aligol@umich.edu**

Hn250@sussex.ac.uk***

mmanthouri@shahed.ac.ir****

**Abstract:**

This study introduces a novel methodology for fault detection and cause identification within the Tennessee Eastman Process (TEP) by integrating a Bidirectional Long Short-Term Memory (BiLSTM) neural network with an Integrated Attention Mechanism (IAM). The IAM combines the strengths of scaled dot product attention, residual attention, and dynamic attention to capture intricate patterns and dependencies crucial for TEP fault detection. Initially, the attention mechanism extracts important features from the input data, enhancing the model's interpretability and relevance. The BiLSTM network processes these features bidirectionally to capture long-range dependencies, and the IAM further refines the output, leading to improved fault detection results. Simulation results demonstrate the efficacy of this approach, showcasing superior performance in accuracy, false alarm rate, and misclassification rate compared to existing methods. This methodology provides a robust and interpretable solution for fault detection and diagnosis in the TEP, highlighting its potential for industrial applications.

## 1) Introduction

Fault diagnosis and prognosis are critical components of industrial process control, as they enable the early detection and prediction of equipment failures, process disturbances, and other anomalies that can impact productivity, safety, and profitability. In industrial applications, fault diagnosis involves the identification of the root cause of a fault or anomaly, while prognosis involves the prediction of the remaining useful life of equipment or the time to failure. Effective fault diagnosis and prognosis can help prevent unexpected shutdowns, reduce maintenance costs, and improve overall process efficiency. However, achieving accurate and reliable fault diagnosis and prognosis in industrial applications is a challenging task, due to the complexity of modern industrial processes, the presence of noise and uncertainty in sensor data, and the need for real-time decision-making [1]. The Tennessee Eastman Process (TEP) dataset, a widely used benchmark dataset in the field of process control and fault diagnosis, has been used to evaluate the performance of various fault diagnosis and prognosis methods, including machine learning algorithms, model-based approaches, and hybrid methods. The results of these studies have demonstrated the potential of advanced fault diagnosis and prognosis techniques to improve process efficiency, reduce maintenance costs, and enhance overall process reliability in industrial applications [2].

LSTM networks, a specialized form of recurrent neural networks, have become prominent in time series prediction and classification due to their effectiveness across diverse domains. Their unique ability to recognize temporal dependencies and complex patterns within sequential data, facilitated by memory cells in their architecture, proves valuable for modeling time-dependent phenomena [3]. LSTM networks have proven to be valuable tools in various applications, including industrial machinery fault detection, predictive maintenance for wind turbines, anomaly detection in network traffic, fault diagnosis in electrical systems, and predictive maintenance for aircraft engines [4], [5]. BiLSTM can capture both short-term and long-term dependencies in time series data, while LSTM can only capture long-term dependencies. This makes BiLSTM better suited for tasks that require understanding both short-term and long-term patterns, such as natural language processing, fault detection, time series prediction [6]. The use of these recurrent neural network architectures has enabled the development of more accurate and reliable fault detection systems, which can significantly reduce downtime, improve maintenance scheduling, and enhance overall system reliability. Furthermore, the ability of LSTM and BiLSTM models to handle large amounts of data and adapt to changing system conditions has made them an attractive solution for real-time fault detection and predictive maintenance applications [7].

The advent of attention mechanisms has revolutionized the field of time series analysis and dynamic data modeling, enabling researchers and practitioners to unlock new insights and improve predictive capabilities [8]. attention mechanisms, inspired by the human brain's ability to focus on relevant information, have introduced a paradigm shift in the way we process and analyze sequential data. By selectively weighting and aggregating input features, attention mechanisms allow models to dynamically adapt to changing patterns and context, effectively "focusing" on the most informative segments of the data [9]. This has led to significant improvements in tasks such as forecasting, anomaly detection, and imputation, particularly in domains characterized by high dimensionality, non-stationarity, and noisy data. Moreover, attention-based models have been shown to be more interpretable and transparent, providing valuable insights into the underlying dynamics of the data and facilitating the identification of key drivers and relationships [10]. The impact of attention mechanisms has been felt across various fields, including finance, healthcare, climate science, and transportation, where the ability to accurately model and predict complex temporal phenomena has far-reaching implications for decision-making, risk management, and policy development treniot [11]. As the complexity and volume of time series data continue to grow, the role of attention mechanisms in unlocking new possibilities for dynamic data modeling and analysis is likely to become even more pronounced, driving innovation and breakthroughs in a wide range of applications.

In this study, a novel fault detection and cause identification methodology for the TEP is presented, employing a combination of a BiLSTM network and a novel Integrated Attention Mechanism (IAM). The innovations are emphasized as follows:

- Feature Extraction with Attention Mechanism: Initially, the attention mechanism is employed to extract important features, discerning their significance in the fault detection process. By calculating attention weights, this approach identifies features that exert a substantial effect on the model's decision-making, enhancing the interpretability and relevance of the selected features for fault detection. Greater attention weights suggest increased importance, pointing to features that play a significant role in influencing the model's output.

- Proposed IAM: This approach integrates the strengths of scaled dot product attention, residual attention, and dynamic attention mechanism to create a comprehensive and adaptive attention mechanism. By leveraging the simplicity and relevance computation of dot product attention, addressing vanishing gradients with the residual attention's residual connection, and incorporating the adaptability of dynamic attention to input sequences. The IAM aims to provide a well-rounded solution that captures nuanced patterns and dependencies in diverse tasks.

- The Significance of Attention Mechanism in Post-BiLSTM Output for Fault Detection: The output generated by the BiLSTM, enriched by the attention mechanism, serves as a refined representation of the input sequence. This post-LSTM output, with enhanced contextual understanding and salient features highlighted by attention, contributes to improved fault detection results. The collaboration between attention mechanisms and BiLSTM optimizes the model's ability to capture intricate patterns, enhancing overall performance in fault detection applications.

The simulation results validate the superior performance and effectiveness of the proposed control strategy compared to existing approaches [12] and [13].

This paper is organized into five sections. The first section provides an overview of the TEP dataset and a review of relevant literature. The next section describes the proposed methodology, which integrates BiLSTM architecture with attention mechanisms. The simulation results are presented in the third section, including the training and evaluation of the BiLSTM-attention hybrid model using the TEP dataset. Finally, the paper concludes with a summary of the main findings and contributions in the last section.

## 2) Literature Review and Dataset Overview

Fault detection is a crucial task in various industrial processes, as it enables the timely identification and mitigation of anomalies, thereby preventing equipment damage, reducing downtime, and improving overall system reliability. The Tennessee Eastman Process (TEP) dataset, a widely-used benchmark for fault detection, provides a realistic simulation of a chemical processing plant, allowing researchers to develop and evaluate fault detection methods in a controlled environment. This section provides an overview of the existing literature on fault detection and introduces the TEP dataset, laying the foundation for the proposed methodology.

## 2.1 Literature Survey of Fault detection

Machine learning has numerous applications across diverse industries, including image and speech recognition in computer vision [14], [15], virtual reality [16] and [17], civil engineering [18] and [19], trajectory prediction [20], medical rehabilitation [21], asset management [22], conversational agents [23], stock prediction [24] and [25], gold price prediction [26], and vehicle routing [27] among many others. As technological landscapes evolve, incorporating cutting-edge methodologies, such as artificial intelligence or machine learning, becomes increasingly relevant for and enhancing anomaly detection and fault diagnosis capabilities [28]. [29] proposes Starnet, a novel framework for ensuring robust edge autonomy by detecting sensor anomalies and evaluating trustworthiness. [30] explores four different techniques for fault detection and isolation in the TEP using principal component analysis, kernel fisher discriminant analysis, and sequential quadratic programming. In [31], a fault detection approach is proposed for chemical processes using independent radial basis function models. [32] proposes a class-incremental fisher discriminant analysis scheme that utilizes a partial F-values with the cumulative per-cent variation to address the limitations of traditional fault detection schemes.

Deep learning methods have emerged as a revolutionary breakthrough in fault diagnosis, outperforming traditional approaches in significant ways. By leveraging complex neural network architectures, deep learning algorithms can learn intricate patterns and relationships in data, enabling them to detect subtle anomalies and faults that may elude traditional methods. The authors in [12] suggest employing a hierarchical deep LSTM supervised autoencoder for the identification and categorization of faults in industrial facilities. [33] This study examines the effectiveness of a robust anomaly detection approach, namely LSTM-based anomaly detection, in identifying unusual patterns in univariate time series data, with a focus on fault detection applications. [34] This paper presents a comparative study of fault diagnosis in the TEP using both LSTM and Back Propagation (BP) models, demonstrating the superiority of LSTM-RNN in terms of accuracy and robustness. [35] proposes a modified BiLSTM for fault diagnosis, which incorporates attention mechanisms and convolutional neural networks to improve the accuracy and robustness of fault detection and classification in complex industrial processes. [36] presents a soft sensor model for the Tennessee Eastman process, which combines a CNN and BiLSTM network with harmony search algorithm to predict key process variables and detect faults with high accuracy.

The attention mechanism has been successfully applied in fault detection applications, enabling models to selectively focus on relevant features and patterns in data, thereby improving the accuracy and robustness of fault detection and classification. [37] presents a novel approach to dynamic gesture recognition using a parallel temporal feature selection framework and an improved attention mechanism. This paper [13] suggest a deep model to employs a seq2seq architecture with attention mechanisms to capture sequential dependencies and identify the root causes of faults of the TEP dataset. [38] proposes the application of a Generalized Transformer model for fault diagnosis in the TEP, leveraging its self-attention mechanism to effectively capture complex relationships between process variables and improve the accuracy of fault detection and classification. [39] presents a self-attention mechanism-based approach for dynamic fault diagnosis and classification in TEP, which uses a novel attention-based neural network to identify and classify faults in real-time, achieving improved accuracy and robustness in fault detection and diagnosis. [40] proposes an improved convolutional neural network architecture that incorporates a multi-head attention mechanism to enhance the fault classification performance in industrial processes, achieving improved accuracy and robustness in identifying faults and anomalies. [41]

presents a novel Twin Transformer architecture that utilizes a Gated Dynamic Learnable Attention (GDLA) mechanism for fault detection and diagnosis in the Tennessee Eastman Process, which enables the model to selectively focus on relevant features and improve the accuracy and robustness of fault detection and classification. Based on the reviewed papers, the attention mechanism has been identified as a novel and efficient approach to fault detection, offering improved accuracy and robustness in identifying faults and anomalies.

**2.1 Tennessee Eastman Process (TEP) dataset**

The Tennessee Eastman Process (TEP) dataset is a widely used benchmark dataset in the field of process control and fault diagnosis. The dataset was created by the Eastman Chemical Company in the 1980s to simulate a realistic industrial process control problem. The TEP dataset simulates a chemical process that involves the production of a hypothetical product, G, from four reactants, A, C, D, and E. The process consists of five major units: a reactor, a condenser, a separator, a stripper, and a recycle stream. The reactor is a continuous stirred-tank reactor (CSTR) where the reactants are mixed and reacted to form the product. The condenser is used to condense the vapor stream from the reactor, and the separator is used to separate the liquid and vapor streams. The stripper is used to remove impurities from the product, and the recycle stream is used to recycle unreacted reactants back to the reactor. The dataset includes 52 measured variables, including temperatures, pressures, flow rates, and compositions, which are sampled every 3 minutes. These variables are used to monitor and control the process, and to detect and diagnose faults [42].

The TEP dataset is particularly useful for testing and evaluating process control and fault diagnosis algorithms because it includes 21 pre-defined fault scenarios, which simulate common process faults such as valve failures, sensor faults, and process disturbances. Each fault scenario is designed to test a specific aspect of process control and fault diagnosis, such as fault detection, isolation, and identification. For example, Fault 1 simulates a step change in the A/C feed ratio, which can cause a disturbance in the process. Fault 2 simulates a sticking valve in the condenser, which can cause a change in the condenser pressure. Fault 3 simulates a sensor failure in the reactor temperature measurement, which can cause a false reading of the reactor temperature. The fault scenarios are designed to be challenging, but not impossible, to detect and diagnose, making the TEP dataset a useful tool for evaluating the performance of process control and fault diagnosis algorithms. The dataset has been widely used in academic and industrial research to evaluate the performance of various process control and fault diagnosis methods, including statistical process control, machine learning, and model-based approaches [43].

The TEP dataset has been widely used in various research studies to evaluate the performance of different process control and fault diagnosis methods. For example, researchers have used the TEP dataset to evaluate the performance of machine learning algorithms, such as neural networks and support vector machines, for fault detection and diagnosis. Other researchers have used the TEP dataset to evaluate the performance of model-based approaches, such as model predictive control and state estimation, for process control and fault diagnosis. The dataset has also been used to evaluate the performance of statistical process control methods, such as control charts and statistical process monitoring, for fault detection and diagnosis. The TEP dataset is available for download from various online sources, and it has become a standard benchmark for evaluating the performance of process control and fault diagnosis algorithms. The dataset is widely used in academic and industrial research, and it has been cited in numerous research papers and publications. The TEP dataset is a valuable resource for researchers and practitioners in the field of process control and fault diagnosis, and it continues to be widely used and studied today.

3)   **Model Description**

In this section, we explore a suite of attention mechanisms, including the foundational scaled dot-product attention, the dynamic attention, and the residual attention mechanism. We introduce the novel proposed IAM, which ingeniously merges these three attention mechanisms, leveraging their unique strengths. The section also delves into the BiLSTM as a robust temporal modeling tool. Our envisioned IAM with BiLSTM forms a seamless fusion, featuring a strategic application of IAM to the input for the initial extraction of dominance fault features. Subsequently, IAM is strategically applied to BiLSTM outputs, fostering a synergistic relationship between attentional capabilities and temporal context modeling. This synergized approach presents a comprehensive strategy for the analysis of sequential data.

**3.1 Scaled Dot-Product Attention Mechanism**

The scaled dot-product attention mechanism, a pivotal component in transformer architectures, is elucidated in this scientific exposition. The neural network module includes a mechanism that adjusts itself during training to prevent problems with the gradients. Through some matrix operations, the attention scores are computed, representing the contextual relevance between query and key vectors. The attention weights, calculated using a Softmax activation function, show how important different parts of the input are. The application of these attention weights facilitates the synthesis of an attended input from the provided value vectors. The scaled dot-product attention mechanism can be expressed mathematically as follows.

Let $Q$, $K$, and $V$ be the query, key, and value matrices, respectively, with dimensions $d_q \times L$, $d_k \times L$, and $d_v \times L$, $d_q$, $d_k$ and $d_v$ represent the dimensions of the query, key, and value vectors, and $L$ denotes the sequence length. The scaling factor is denoted by $\sqrt{d_k}$. The term $\sqrt{d_k}$ represents the square root of the dimension $d_k$. In the context of attention mechanisms, it is commonly used as a scaling factor. Note that, in many cases, especially in self-attention mechanisms like those used in transformer models, $Q$, $K$, and $V$ are set to be the same, coming from the same input sequence; hence in this study, they are assumed to be equal to the input sequence (x). The attention scores $A$ are computed as follows.

$$A = Softmax\left(\frac{QK^T}{\sqrt{d_k}}\right). \tag{1}$$

The attended input $Z$ is then obtained by multiplying the attention scores with the value matrix:

$$Z = A \times V, \tag{2}$$

this formulation encapsulates the essence of the scaled dot-product attention mechanism, providing a concise representation of its underlying mathematical operations within the context of transformer-based neural network architectures. Finally, (3) represents the calculation of attended input in the scaled dot-product attention mechanism.

$$Z(Q,K,V) = Softmax\left(\frac{QK^T}{\sqrt{d_k}}\right) \times V. \tag{3}$$

The block diagram of the scaled dot-product attention mechanism is shown in Fig. 2.

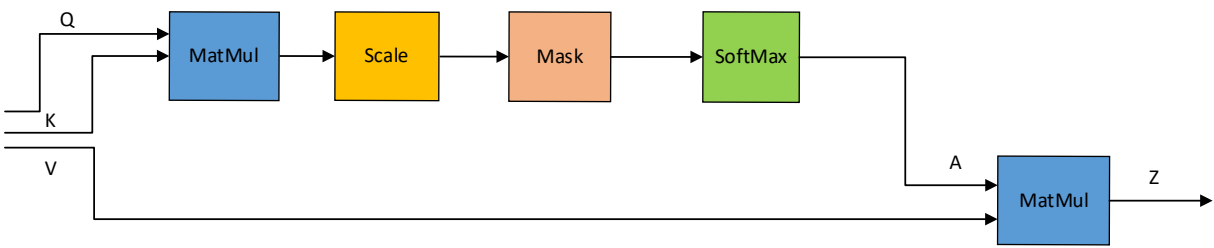

Fig. 2. The block diagram of the scaled dot-product attention mechanism

### 3.2 Dynamic Scaled Dot-Product Attention Mechanism

The dynamic scaled dot-product attention mechanism introduces an innovative dimension to the scaled dot-product attention mechanism by incorporating an adjustable scaling parameter.

This dynamic feature enables the model to dynamically regulate the influence of attention scores, offering a tunable mechanism for adapting to diverse data patterns and complexities. The attention scores in the dynamic scaling $A$ are computed as

$$scores = \frac{QK^T}{\sqrt{d_k}}, \tag{4}$$

$$dynamic\_scores = scores \odot \lambda, \tag{5}$$

$$A = Softmax(dynamic\_scores), \tag{6}$$

where $\lambda$ is the additional adaptive scaling parameter and $\odot$ is the sign of element-wise multiplication. Equ. (7) shows the attended input $Z$ is then obtained by multiplying the attention scores with the value matrix.

$$Z = A \times V, \tag{7}$$

in this formulation, the introduction of an additional adaptive scaling parameter $\lambda$ offers a dynamic mechanism to finely tune the impact of attention scores on the final weights. $\lambda$ is a flexible parameter for the attention mechanism thereby bolstering its adaptability and expressiveness in capturing intricate relationships within the input data. Hence, the overall formula is equal to (8).

$$Z_{Dynamic}(Q, K, V) = Softmax\left(\frac{QK^T}{\sqrt{d_k}} \odot \lambda\right) \times V. \tag{8}$$

The block diagram of the dynamic scaled dot-product attention mechanism is illustrated in Fig. 3.

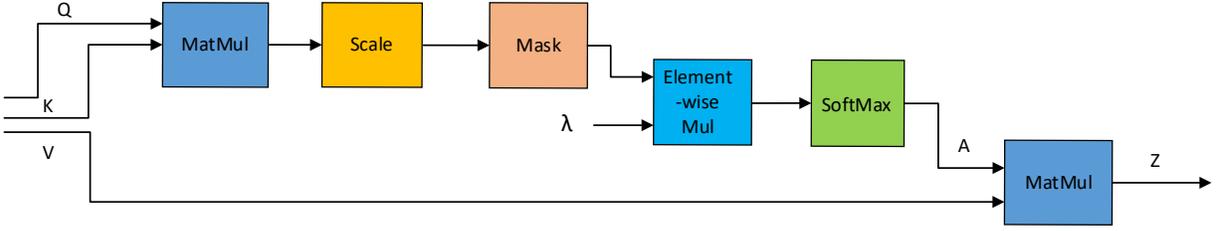

Fig. 3. The block diagram of the dynamic scaled dot-product attention mechanism

**3.3 Residual Attention Mechanism**

Residual attention is a sophisticated extension of traditional attention mechanisms within neural network architectures. In this paradigm, the residual connections, inspired by the success of residual networks, are seamlessly integrated with attention mechanisms to enhance information flow and gradient propagation during training. The key idea is to create shortcut connections that bypass certain attention layers, allowing the network to directly access and preserve the original input information. This mitigates the risk of information loss during the attention process, addressing challenges such as vanishing gradients and facilitating the training of deeper models. By combining the strengths of residual connections and attention mechanisms, residual attention achieves a delicate balance between capturing intricate patterns in the data and maintaining the integrity of the input features, leading to more effective and stable learning in complex tasks. This approach has demonstrated notable success in various applications, showcasing its potential to improve the performance and convergence of deep neural networks [44]. Given an input tensor $x$ with a transformation function of attention mechanism $F(x)$, the output $y$ of the residual block is calculated as

$$y = x + F(x), \qquad (9)$$

this formulation embodies the essence of residual connections, where the original input is preserved through a direct shortcut connection, and the attention-based transformation $F(x)$ is added to it. The residual block architecture allows the model to learn residual features, capturing both the original information and the refined features generated by the attention mechanism. This not only aids in the prevention of information loss but also facilitates smoother gradient flow during training, contributing to the stability and effectiveness of deep neural networks.

### 3.4 The Proposed Integrated Attention Mechanism

We integrated the dynamic scaled dot-product attention with the residual attention mechanism, creating a unified approach known as the IAM. The final formulation is expressed as:

$$Y = x + Softmax\left(\frac{QK^T}{\sqrt{d_k}} \odot \lambda\right) \times V = x + Z_{Dynamic}(Q, K, V), \tag{10}$$

in this formulation, the original input $x$ is combined with the attended input obtained through the dynamic scaled dot-product and residual attention mechanisms, providing the model with the ability to selectively emphasize relevant features while preserving the underlying information. Fig. 4 shows the IAM block diagram.

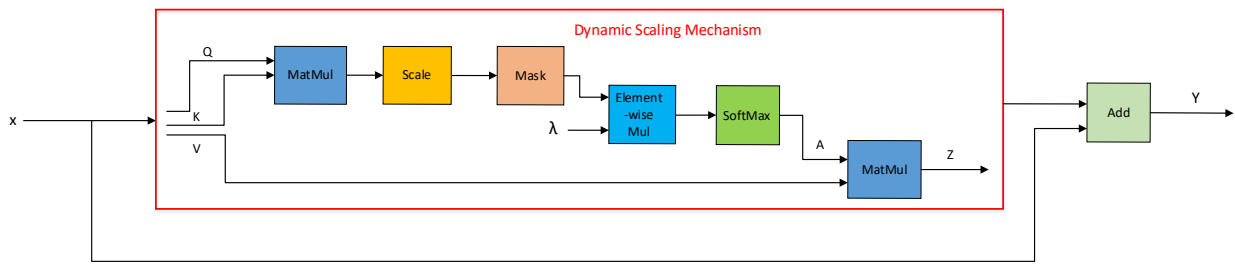

Fig. 4. The block diagram of the proposed IAM

### 3.5 Bidirectional Long Short-Term Memory (BiLSTM)

LSTM networks have revolutionized the field of recurrent neural networks (RNNs) by providing an elegant solution to the problem of learning and capturing long-term dependencies in sequential data. Unlike traditional RNNs, which suffer from the vanishing gradient problem, LSTMs are specifically designed to overcome this limitation. They achieve this by employing specialized units, such as forget gates, input gates, and output gates, which are equipped with adaptive mechanisms that enable the network to selectively retain or discard information over various time scales. The LSTM layer can be represented by the following equations.

$$f_t = \sigma(W_f \cdot [h_{t-1}, x_t] + b_f), \tag{11}$$

$$i_t = \sigma(W_i \cdot [h_{t-1}, x_t] + b_i), \tag{12}$$

$$\tilde{C}_t = tanh(W_C \cdot [h_{t-1}, x_t] + b_C), \tag{13}$$

$$C_t = f_t * C_{t-1} + i_t * \tilde{C}_t, \tag{14}$$

$$o_t = \sigma(W_o \cdot [h_{t-1}, x_t] + b_o), \tag{15}$$

$$h_t = o_t * tanh(C_t), \tag{16}$$

where $f_t$, $i_t$, $o_t$ are the forget, input, and output gates, respectively; $C_t$ is the cell state; $h_t$ is the hidden state; $x_t$ is the input at time step $t$; $\sigma$ is the sigmoid function; and $W$ and $b$ are weight matrices and bias vectors, respectively.

BiLSTM is a variant of the traditional LSTM recurrent neural network architecture. The primary distinction lies in its ability to process input sequences in both forward and backward directions, enabling the model to capture contextual information from both past and future time steps. Mathematically, the output of a BiLSTM at a given time step $t$ can be represented as:

$$h_t^{(BiLSTM)} = [\vec{h}_t, \overleftarrow{h}_t],$$

where $\vec{h}_t$ is the hidden state of the forward LSTM at time step $t$ and $\overleftarrow{h}_t$ is the hidden state of the backward LSTM at the same time step. The brackets represent the concatenation operation. This bidirectional processing enhances the model's ability to capture long-range dependencies and context in sequential data, making BiLSTM a popular choice for tasks such as natural language processing, speech recognition, and time series prediction.

**3.6 The Proposed IAM with BiLSTM**

This subsection introduces a designed sequential processing method that synergizes the power of the IAM and BiLSTM neural networks for fault detection in sequential data. The orchestrated sequence of operations unfolds as follows.

1. IAM applied to input:

- The output of the IAM applied to the input is given by:

$$Y_{t_1} = x_t + Softmax\left(\frac{QK^T}{\sqrt{d_k}} \odot \lambda\right) \times V = x_t + Z_{Dynamic}(Q_t, K_t, V_t), \tag{17}$$

the IAM selectively attends to relevant features in the input, generating an intermediate representation denoted as $Y_{t_1}$.

2. BiLSTM Output of IAM:

- The BiLSTM is applied to the output of the IAM:

$$h_{t_1} = BiLSTM(Y_{t_1}). \tag{18}$$

$Y_{t_1}$ is then fed into a BiLSTM layer, producing $h_{t_1}$. This step captures temporal dependencies and refines the feature representation.

3. IAM Applied to BiLSTM Output:

- The IAM is again applied to the output of the BiLSTM:

$$Y_{t_2} = h_{t_1} + Z_{Dynamic}\left(Q_{h_{t_1}}, K_{h_{t_1}}, V_{h_{t_1}}\right), \tag{19}$$

the IAM is applied once again to $h_{t_1}$, yielding $Y_{t_2}$. This dual application enhances the model's ability to focus on crucial aspects of temporal context, considering the refined features generated by the BiLSTM.

4. Two Fully Connected Layers:

- The output $Y_{t_2}$ is then processed through two fully connected layers:

$$Y_3 = FC_1(Y_2), \tag{20}$$

$$output = FC_2(Y_3), \tag{21}$$

the output $Y_{t_2}$ undergoes processing through two fully connected layers ($FC_1$ and $FC_2$), resulting in the final output.

In this method, the sequential processing unfolds in a carefully orchestrated manner to capture intricate patterns and dependencies in the input data. First, the IAM is applied to the input, producing an intermediate representation denoted as $Y_{t_1}$. This step involves selectively attending to relevant features through a dynamic attention mechanism, as expressed in (17). Subsequently, the output $Y_{t_1}$ is fed into BiLSTM layer, denoted as $h_{t_1}$, which captures temporal dependencies and refines the feature representation. The IAM is applied once again to the output of the BiLSTM, yielding $Y_{t_2}$. This dual application of attention mechanisms enhances the model's ability to focus on crucial aspects of the temporal context while considering the refined features generated by the BiLSTM. Following this, the $Y_{t_2}$ is processed through two FC layers $FC_1$ and $FC_2$, providing the final output. This sequence of operations allows the model to learn and extract hierarchical features, capturing both spatial and temporal information effectively. The proposed fault detection strategy integrates IAM as a feature extractor, demonstrating its versatility and adaptability. The IAM's application to both the input and BiLSTM output signifies a holistic approach, leveraging attentional capabilities and temporal context modeling for improved fault detection performance.

The architecture, illustrated in Fig. 5, visually represents this method's comprehensive strategy, showcasing the sequential flow of operations and the integration of IAM with BiLSTM and FC layers.

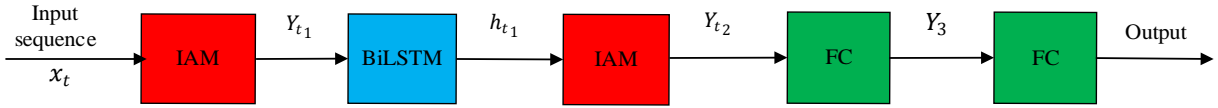

Fig. 5. The block diagram of the proposed IAM with BiLSTM and FC layers.

In this study, we propose a neural network architecture tailored for sequence-based data, incorporating BiLSTM layers and fully connected layers to ensure a balance between computational efficiency and model expressiveness. The architecture details and training parameters are meticulously chosen to optimize performance. The input to the model consists of sequences with the following characteristics:

- Batch size: 64
- Sequence length: 10
- Number of features: 52

Hence $K, Q, V \in \mathbb{R}^{32 \times 10 \times 52}$. The hidden size of the BiLSTM is 128, and the architecture includes two fully connected layers with 128 and 64 neurons, respectively, each with a 0.2 dropout rate and

ReLU activation function. The Adam optimizer is employed with an initial learning rate of 0.001, which adaptively reduces to 1e-4. The model is trained for a total of 200 epochs.

4) **Simulations**

In this section, we delve into the training and testing phases of our proposed IAM with BiLSTM for fault diagnosis, showcasing its capabilities and highlighting the importance of attention weights in feature selection. Our evaluation goes beyond traditional metrics, incorporating precision, recall, F1-score, false discovery rate, false alarm rate, and misclassification rate for each fault class, providing a comprehensive understanding of the model's performance across various fault categories. We also compare our results with those of existing state-of-the-art methods, offering a nuanced assessment of our model's effectiveness.

**4.1 Accuracy**

In Figs. 6 and 7, the graphical depictions illustrate the training and testing metrics over 200 epochs with a fixed learning rate of 0.001 and a batch size of 64. These visual representations offer a comprehensive insight into the model's performance, showcasing how it evolves and generalizes over the course of the training process.

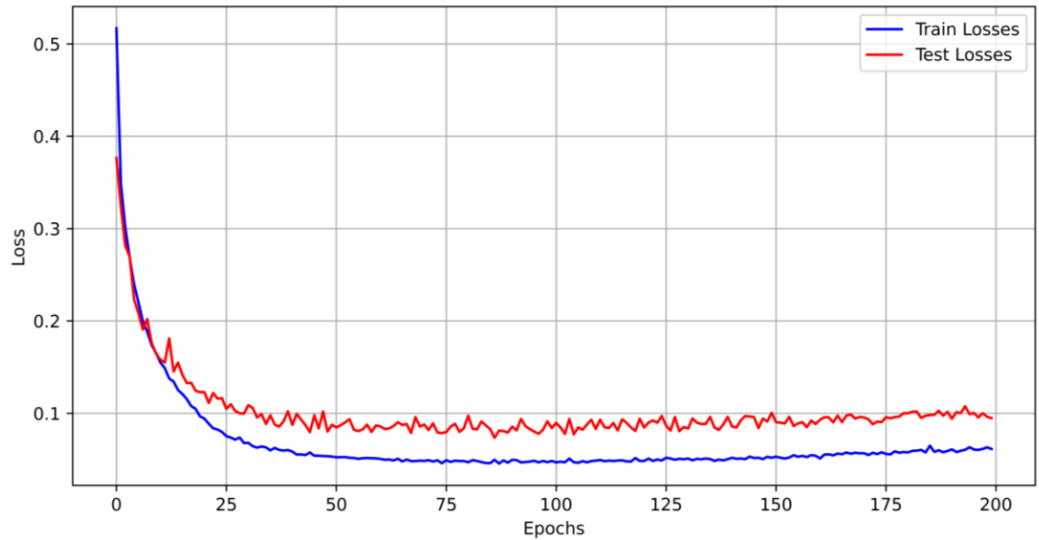

Fig. 6. The train/test losses of the proposed approach during training process

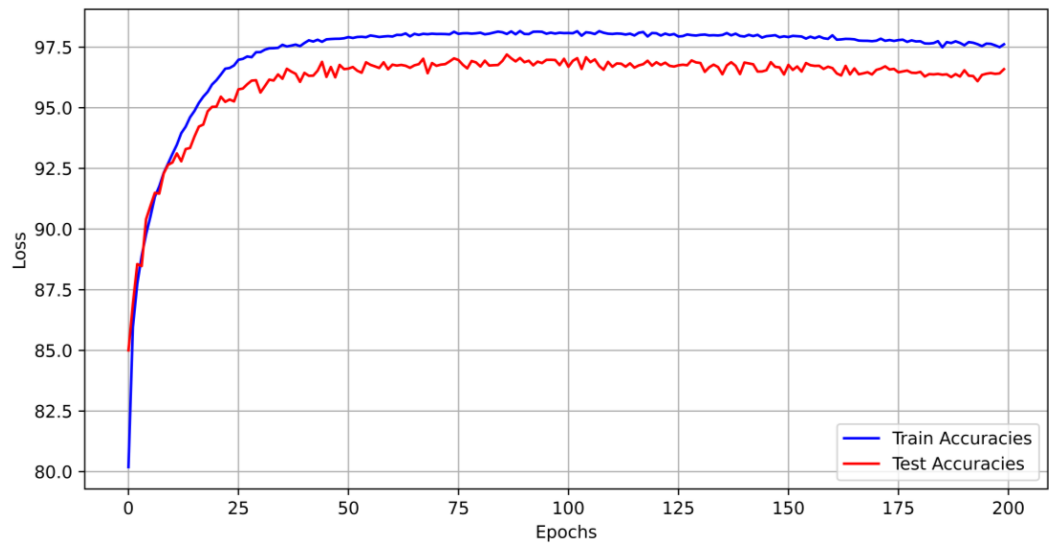

Fig. 7. The train/test accuracies of the proposed approach during training process

Looking at Fig. 6 and Fig. 7, epoch 87 and epoch 108 are especially important points, each one showing that the model reached a major breakthrough in its development over time. Here are the detailed achievements associated with these key epochs.

Minimum Test Loss at Epoch 87:

- Test Loss: 0.0734

- Test Accuracy: 97.19%

- Training Loss: 0.0496

- Training Accuracy: 98.04%

Minimum Training Loss at Epoch 108:

- Training Loss: 0.0469

- Training Accuracy: 98.16%

- Test Loss: 0.0879

- Test Accuracy: 96.72%

At Epoch 87, the model achieved a remarkable balance with a minimum test loss of 0.0734 and a corresponding high-test accuracy of 97.19%. The training loss of 0.0496 and training accuracy of 98.04% underscore the model's proficiency in learning and generalizing from the training data. Epoch 108 further solidified the model's capabilities, reaching a minimum training loss of 0.0469 with an impressive training accuracy of 98.16%. Despite a slight increase in the test loss to 0.0879, the test accuracy remained high at 96.72%.

In both epochs, the model demonstrated a strong foundation for fault detection, showcasing its ability to minimize errors and generalize well to new, unseen data. The nuanced interplay between training and testing metrics reflects the model's robustness and effectiveness in fault diagnosis tasks. Choosing the model from Epoch 87, with the minimum test loss of 0.0734, aligns with the principle of selecting a model that generalizes well to new, unseen data.

## 4.2 Cause Identification

In this section, we investigate the visualization of attention weights, a key element in uncovering the root causes of faults. Leveraging the proposed IAM, we unravel the pivotal features that exert a major influence on fault detection. Fig. 8 depicts the visualization of attention weights, illustrating their role in both normal operation and the identification of causes for faults 1 and 2.

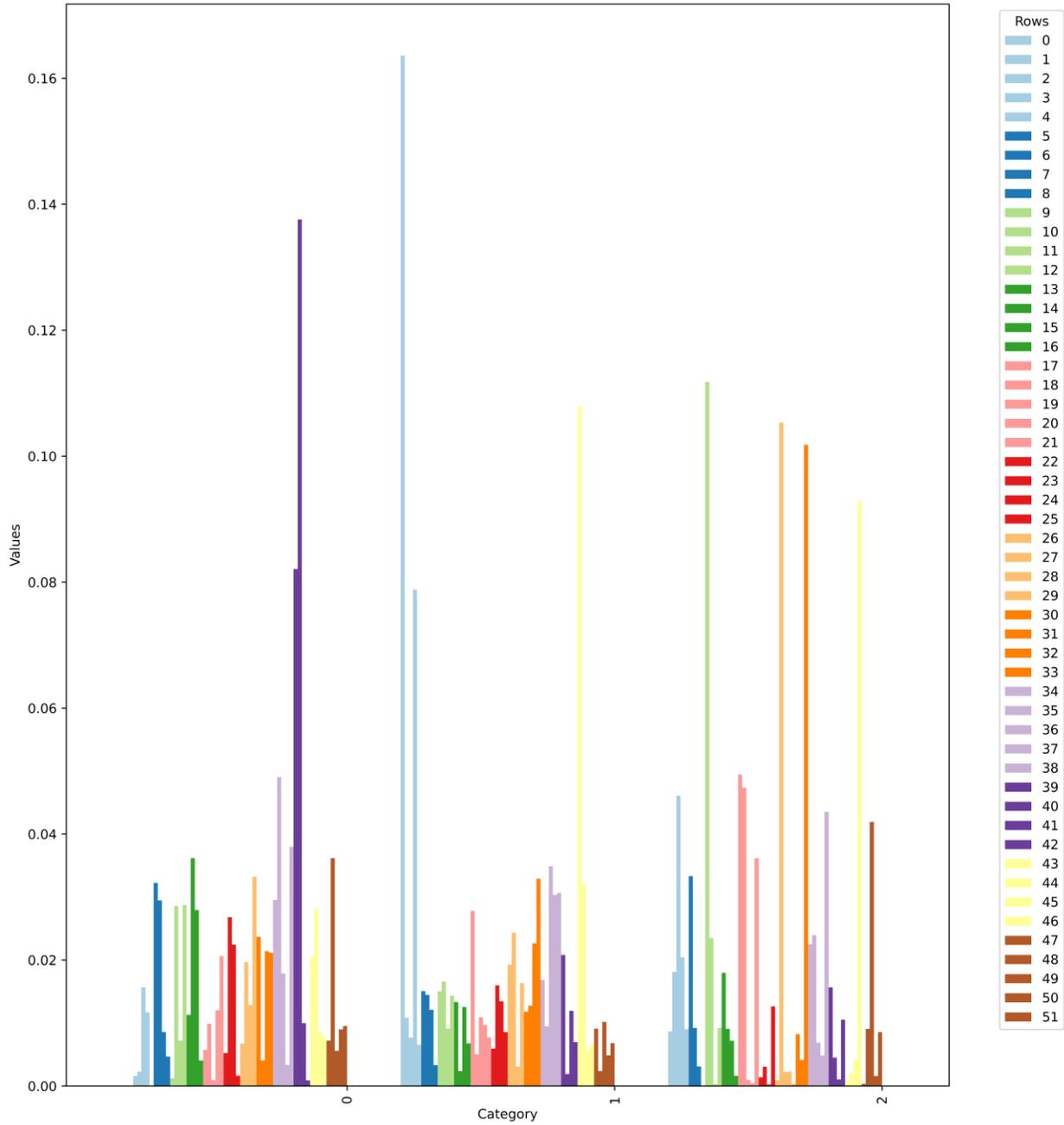

Fig. 8. The attention weights visualization for normal operation and identifying causes of faults 1,2

Analyzing Fig. 8 provides valuable insights into the attention weights visualization for normal operation and the identification of faults 1 and 2. Notably, the visualization highlights the following major influential features.

- Normal Operation:

Features 39 and 40 emerge as significant contributors to the fault detection process during normal operation. The attention weights indicate their pronounced impact in discerning normal data patterns.

- Fault 1:

For fault 1, attention is notably directed towards features 0, 3, and 43. These features play a pivotal role in the model's ability to detect and differentiate fault 1 patterns from normal operation.

- Fault 2:

Fault 2 detection is influenced prominently by features 9, 27, 33, and 46. The attention weights underscore the significance of these features in identifying and isolating fault 2 occurrences.

This detailed analysis not only sheds light on the crucial features for fault detection in each scenario but also enhances our understanding of the model's interpretability and its ability to discern diverse fault patterns. Figures 9 to 14 display the attention weights pertaining to faults 3 through 20.

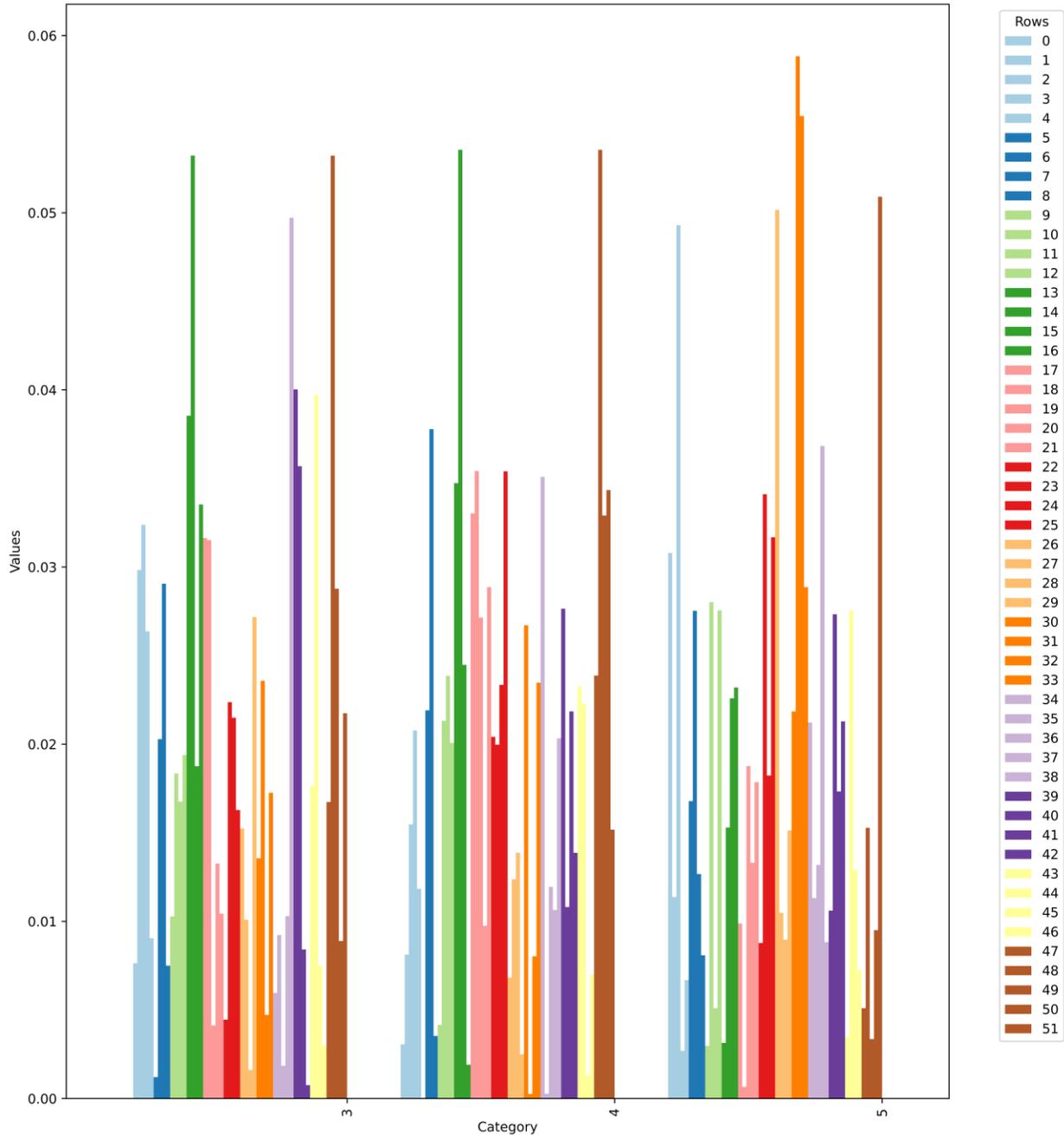

Fig. 9. The attention weights visualization for identifying causes of faults 3, 4, and 5

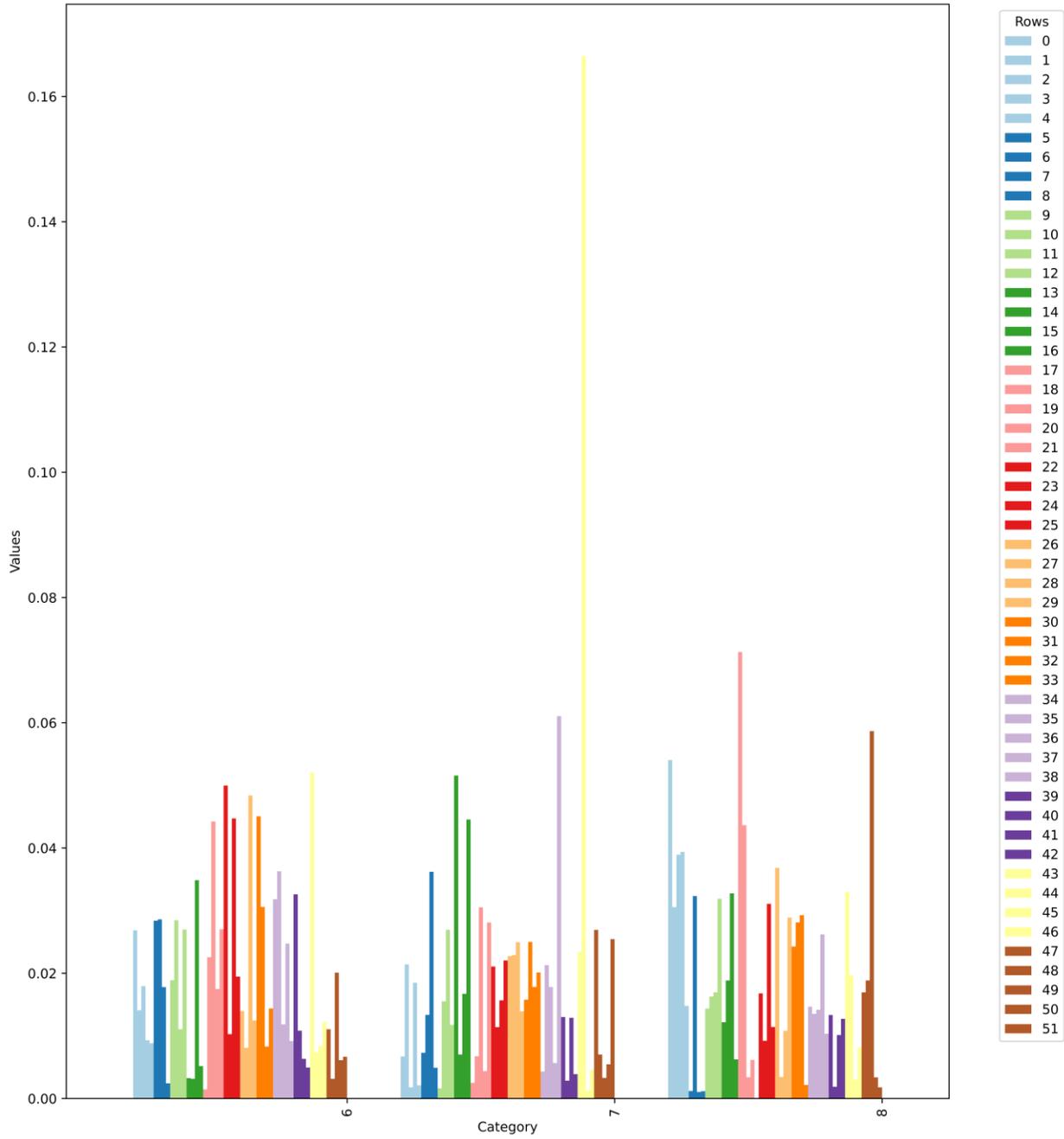

Fig. 10. The attention weights visualization for identifying causes of faults 6, 7, and 8

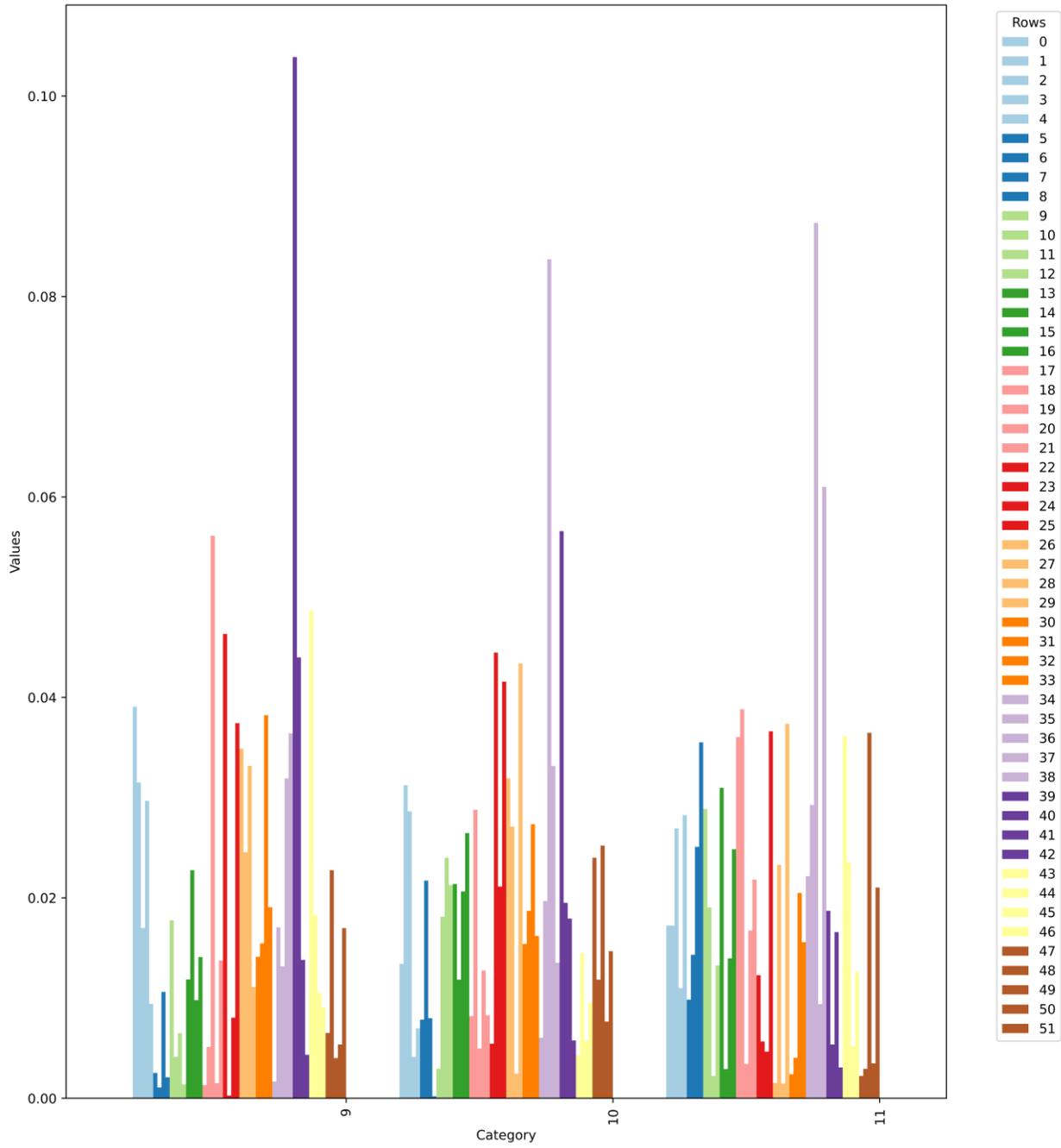

Fig. 11. The attention weights visualization for identifying causes of faults 9, 10, and 11

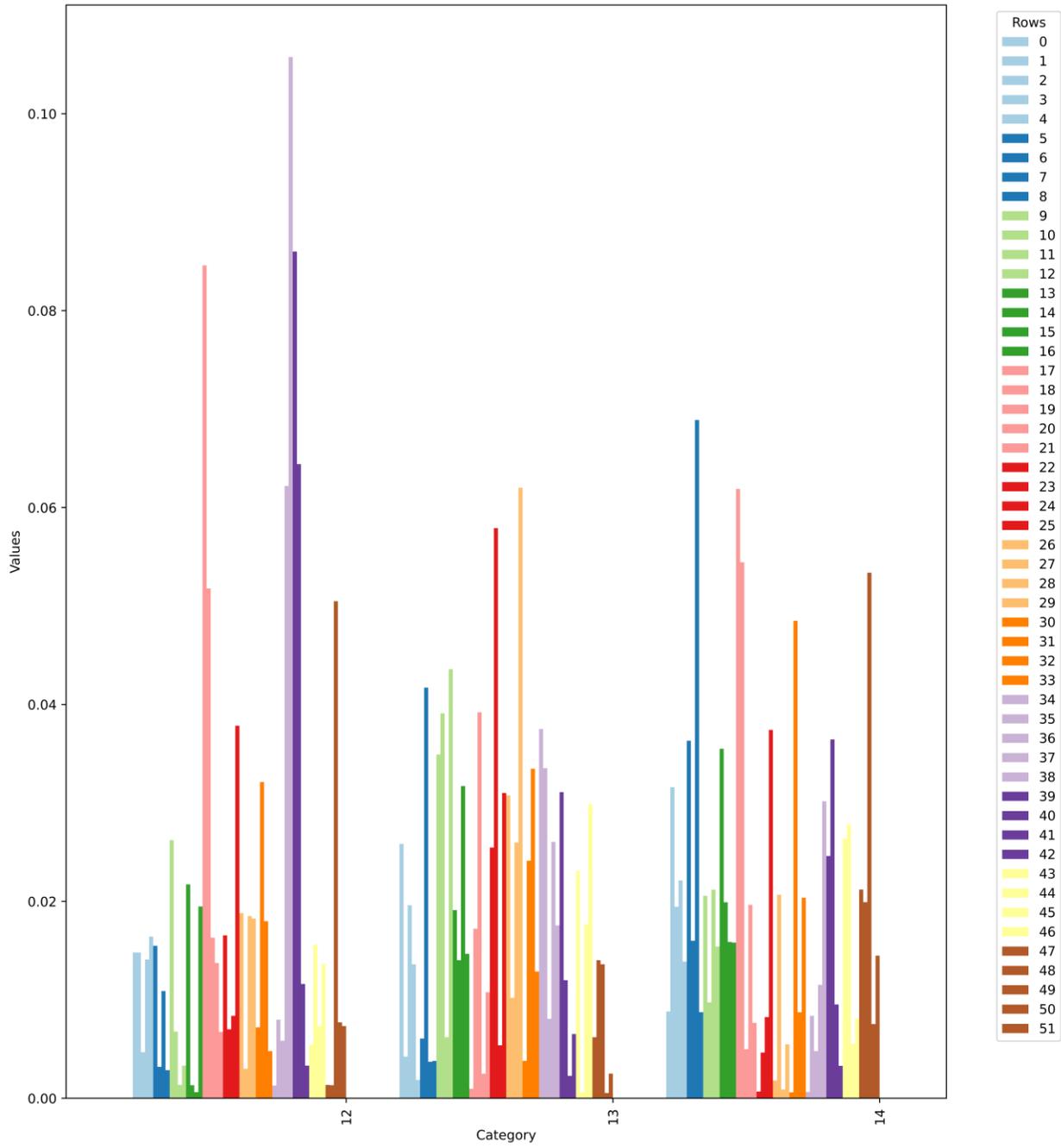

Fig. 12. The attention weights visualization for identifying causes of faults 12, 13, and 14

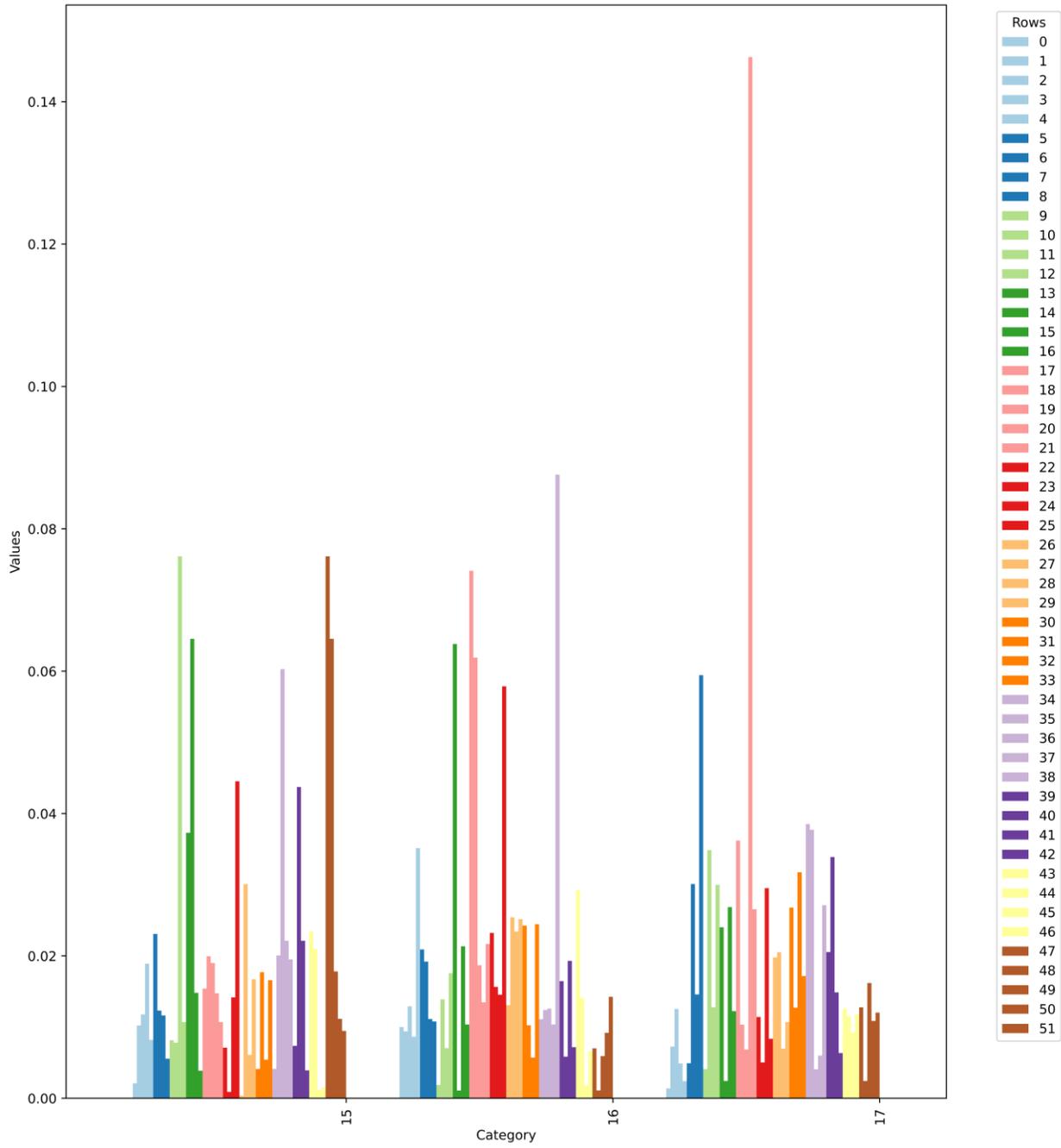

Fig. 13. The attention weights visualization for identifying causes of faults 15, 16, and 17

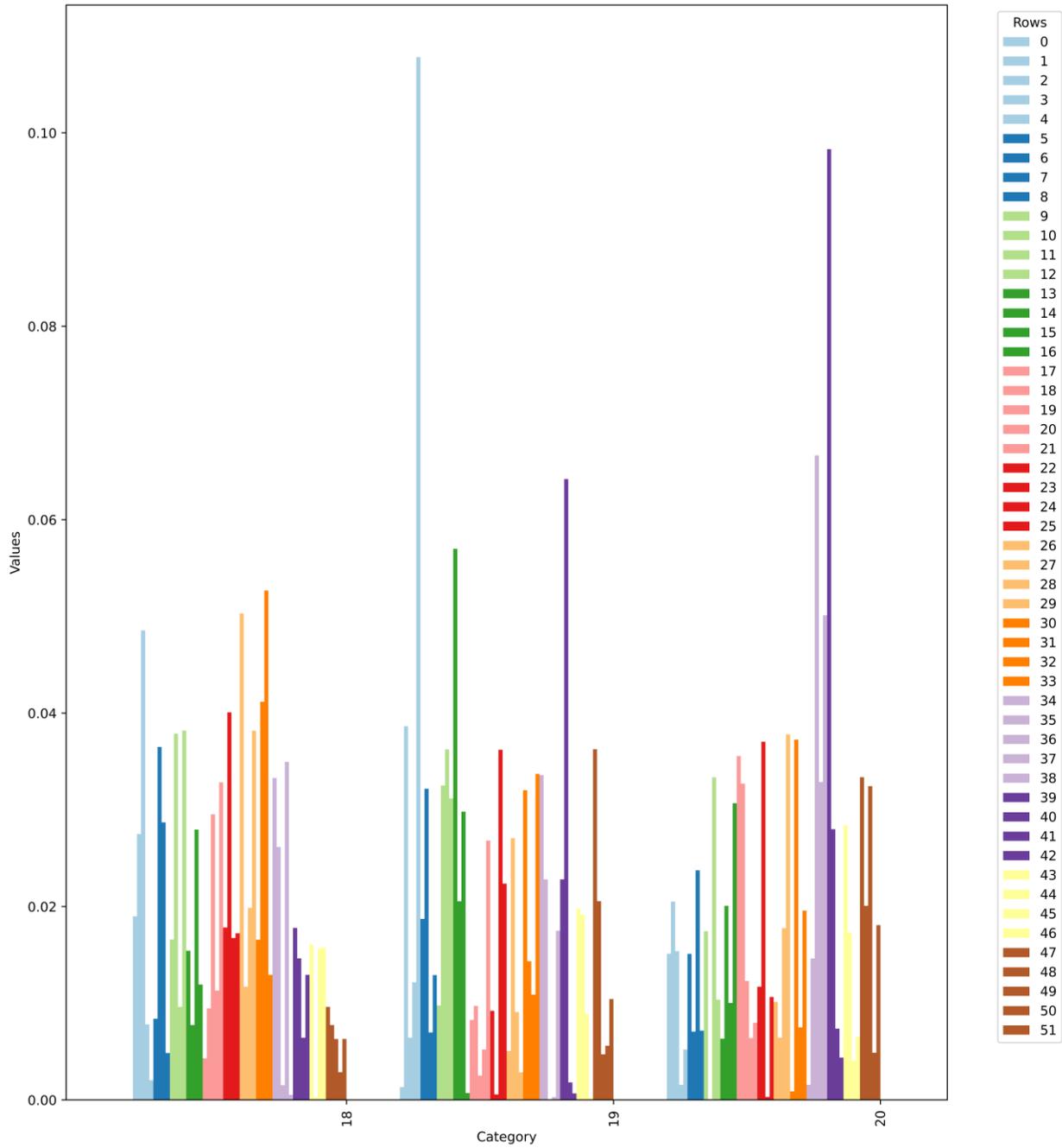

Fig. 14. The attention weights visualization for identifying causes of faults 18, 19, and 20

The importance of features for faults 3 to 20 can be comprehended in a manner similar to the insights gained from Fig. 8.

Overall, the IAM with BiLSTM demonstrates superior fault detection proficiency, especially in terms of F1-score, compared to state-of-the-art methods such as Deep LSTM-SAE [12]. Notably, IAM with BiLSTM achieves this high performance with a lower parameter count, highlighting its efficiency and potential for resource-conscious implementations.

5) **Conclusion**

This study proposes a novel fault detection and cause identification methodology for the TEP using a BiLSTM neural network enhanced with an innovative IAM. The proposed IAM incorporates scale dot product attention, residual attention, and dynamic attention to effectively capture intricate patterns and dependencies within the TEP data. The IAM's unique role in supervised feature extraction distinguishes the proposed methodology. By calculating attention weights, the IAM identifies key features that significantly impact fault detection, providing interpretability and insight into the model's decision-making process. This proposed IAM with BiLSTM method presents a versatile and effective approach for capturing intricate patterns and dependencies crucial for TEP fault detection. Comparative evaluations demonstrate the proposed IAM with BiLSTM approach's superior performance in fault detection, particularly in terms of F1-score, outperforming state-of-the-art methods. The well-balanced performance, evidenced by the equilibrium between capturing positive instances and minimizing false positives, underlines the robustness of the proposed methodology.

Future work on this paper could include exploring other attention mechanisms, applying the methodology to other industrial processes, handling noisy or missing data, scaling and implementing the model in real-time, comparing with other machine learning approaches, investigating the impact of hyperparameters, applying to other types of faults, and developing a more interpretable model.